\documentclass{article} 
\usepackage{iclr2026_conference}
\iclrfinalcopy

\usepackage{amsmath,amsfonts,bm}

\def\eqref#1{equation~\ref{#1}}

\def\1{\bm{1}}

\DeclareMathAlphabet{\mathsfit}{\encodingdefault}{\sfdefault}{m}{sl}
\SetMathAlphabet{\mathsfit}{bold}{\encodingdefault}{\sfdefault}{bx}{n}

\usepackage{hyperref}
\usepackage{url}
\usepackage{subcaption}

\PassOptionsToPackage{numbers, compress}{natbib}
\usepackage{graphicx} 
\usepackage{amsmath} 
\usepackage{algorithm}
\usepackage{comment}
\usepackage{ragged2e}

\usepackage{algpseudocode}

\usepackage[utf8]{inputenc} 
\usepackage[T1]{fontenc}    
\usepackage{hyperref}       
\usepackage{url}  
\usepackage{multirow} 
\usepackage{booktabs}       
\usepackage{amsfonts}       
\usepackage{nicefrac}       
\usepackage{microtype}      
\usepackage{xcolor}         
\usepackage{wrapfig}

\title{Circuit-Aware Reward Training: A Mechanistic Framework for Longtail Robustness in RLHF}

\author{
  Jing Liu \\
  ENS, Université PSL, EHESS, CNRS \\
  Paris, France \\
  \texttt{jing.liu@psl.eu} \\
}

\begin{document}

\maketitle

\begin{abstract}
Reinforcement Learning from Human Feedback (RLHF) reward models exhibit systematic failures on longtail distributions, leading to reward hacking and misalignment. We propose a mechanistic interpretability framework that identifies specialized neural circuits responsible for rare-event processing in reward models. Drawing from recent advances showing distributed specialization for rare tokens in language models\citep{liu2025no, liu2025emergent}, we hypothesize that reward models also develop functionally distinct circuits for longtail scenarios. Our theoretical framework establishes formal connections between circuit specialization, reward generalization bounds, and longtail performance. We introduce \textbf{Circuit-Aware Reward Training (CART)}, which uses circuit analysis to guide data augmentation, regularization, and ensemble strategies. This approach provides both theoretical insights into reward model failures and practical interventions for improving longtail robustness.
\end{abstract}

\section{Introduction}

Reinforcement Learning from Human Feedback (RLHF) has become a cornerstone for aligning large language models (LLMs) with human preferences. However, reward models in RLHF often exhibit \textbf{longtail failures}, struggling to generalize to rare or underrepresented inputs, which can lead to reward hacking, misalignment, and unpredictable behaviors~\citep{lilianweng2024rewardhacking}.

Mechanistic interpretability (MI) offers a promising tool to diagnose and mitigate these issues. Recent studies have revealed that LLMs process rare tokens through \textbf{distributed specialization}: spatially distributed subnetworks that activate specifically for rare inputs, without requiring modular routing mechanisms~\citep{liu2025no}. We hypothesize that reward models exhibit similar specialization, with distinct circuits processing rare preference scenarios. Understanding these circuits provides a new lens for diagnosing and preventing reward hacking.

Existing work has explored various aspects of reward model behavior. Pitis~\citep{pitis2023failure} identified multiple failure modes in reward models, including inconsistencies and biases that degrade model performance. 
We argue that reward hacking is not merely an optimization problem but fundamentally a \emph{longtail generalization failure}. Reward models, trained predominantly on common preference scenarios, develop systematic blind spots for rare edge cases. These blind spots become exploitable vulnerabilities during policy optimization, as models learn to navigate toward underrepresented regions where reward estimates are unreliable.

Despite these advancements, a comprehensive understanding of how reward models generate rare tokens remains limited. We propose leveraging MI to analyze the internal mechanisms of reward models, uncovering specialized subnetworks responsible for rare-event processing. These insights can guide targeted interventions, improving alignment and generalization on longtail scenarios.

This position paper makes three key contributions. First, we provide a novel mechanistic explanation for reward hacking grounded in circuit specialization theory. Second, we formalize the connection between circuit complexity and longtail generalization through theoretical bounds. Third, we introduce \textbf{Circuit-Aware Reward Training (CART)}, a practical methodology that leverages mechanistic insights to improve reward model robustness.

\section{Related Work}

To mitigate reward hacking and longtail failures, several approaches have been proposed. 

\textbf{Reward shaping} methods modify the reward function to penalize undesired behaviors or emphasize success on rare-event inputs. Techniques like WARM and Minmax aim to guide model optimization toward intended behaviors while discouraging exploitation of proxies~\citep{fu2025reward}. However, these methods often rely on heuristics and require extensive domain knowledge to define appropriate shaping signals, limiting scalability.

\textbf{Regularization-based approaches} such as KL-regularization, label smoothing, and hidden state dropout constrain model updates, preventing overfitting to sparse feedback~\citep{yang2024regularizing}. While effective in reducing extreme misalignment, these methods operate at a coarse level and do not explicitly account for internal representations of rare-event inputs, leaving longtail failures largely unaddressed.

\textbf{Pessimistic reward modeling} tackles overestimation in underrepresented regions by encouraging the model to predict conservative reward estimates for rare cases~\citep{xu2025learning}. This approach improves robustness but depends heavily on careful tuning of pessimism parameters and does not illuminate which internal components of the model drive the behavior.

\textbf{Contrastive and ensemble methods} such as RewardFusion and contrastive reward modeling compare candidate outputs and aggregate predictions across multiple reward models~\citep{shen2025rewardfusion}. These techniques reduce reward variance and improve generalization, yet they treat the model as a black box, providing little insight into how rare inputs are represented or processed internally.

\textbf{Information-theoretic approaches} like InfoRM introduce variational bottlenecks to filter irrelevant information, limiting overoptimization on spurious features~\citep{liu2024information}. While promising, these methods focus on input-feature selection rather than uncovering the functional organization of neurons or subnetworks that govern rare-event processing.

Current approaches to reward hacking largely treat it as a black-box optimization problem. Techniques like KL regularization, reward clipping, and ensemble methods address the symptoms without understanding the underlying mechanisms. While these approaches provide some mitigation, they often involve trade-offs between reward optimization and constraint satisfaction.

This motivates our proposal to leverage mechanistic interpretability to uncover \textbf{specialized subnetworks} responsible for longtail behaviors, to enable targeted interventions and more robust RLHF reward models. Specifically, we hypothesize that reward models develop specialized neural circuits for processing different types of preference scenarios, with longtail circuits being particularly vulnerable to exploitation.

\section{Circuit-Aware Reward Training: A Practical Framework}

Just as language models develop specialized circuits for different linguistic phenomena such as induction heads for in-context learning and factual recall circuits for knowledge retrieval\citep{elhage2021mathematical, olah2020zoom, bereska2024mechanistic}, reward models develop distinct circuits for different preference domains. Crucially, circuits responsible for rare preference scenarios (longtail circuits) receive limited training signal, making them prone to overconfident and inconsistent predictions. This vulnerability can be exploited by policies, leading to reward hacking.

\subsection{Circuit Specialization in Reward Models}

To reason about such specialization formally, let $R_\theta: \mathcal{X} \rightarrow \mathbb{R}$ denote a reward model mapping input $x$ to a scalar score. Inspired by mechanistic interpretability work \citep{olah2020zoom, marks2024sparse}, we conceptually decompose its computation into circuit activations:
\begin{equation}
R_\theta(x) = \sum_{i=1}^k w_i \cdot a_i(x) + \epsilon(x),
\end{equation}
where $a_i(x)$ represents the activation of circuit $C_i$, $w_i$ its output weight, and $\epsilon(x)$ residual computation. This decomposition provides a principled way to isolate the contributions of different functional components, allowing us to analyze how longtail circuits may produce systematic errors when undertrained.

We now establish formal connections between circuit specialization and reward generalization. Consider a mixture distribution of common and rare preference scenarios:
\begin{equation}
p(x) = \alpha p_{\text{head}}(x) + (1-\alpha) p_{\text{tail}}(x),
\end{equation}
where $\alpha$ reflects the prevalence of common scenarios. We define a circuit $C_i$ as $\tau$-specialized for the tail if:
\begin{equation}
\mathbb{E}_{x \sim p_{\text{tail}}}[|a_i(x)|] - \mathbb{E}_{x \sim p_{\text{head}}}[|a_i(x)|] > \tau.
\end{equation}

Intuitively, this captures the idea that certain circuits respond preferentially to rare inputs, which aligns with empirical observations of sparse or modular neural mechanisms in LLMs \citep{marks2024sparse, templeton2024scaling,liu2025no}.

\subsection{Circuit Complexity and Generalization}

Under this formalization, we can derive a bound on generalization error for longtail circuits. For a reward model with longtail-specialized circuits $\mathcal{C}_{\text{tail}}$, the generalization error on rare inputs is bounded by:

\begin{align}
\mathcal{L}_{\text{tail}}(\theta) \leq &\mathcal{L}_{\text{head}}(\theta) + \frac{C\sqrt{|\mathcal{C}_{\text{tail}}|\log(1/\delta)}}{\sqrt{N_{\text{tail}}}} \\ 
&+ \beta \cdot \text{Div}(\mathcal{C}_{\text{tail}}, \mathcal{C}_{\text{head}})
\end{align}

where $N_{\text{tail}}$ is the number of longtail training examples, $C$ is a constant, and $\text{Div}(\cdot,\cdot)$ measures circuit divergence.

This bound reveals two key insights: (1) longtail error increases with the number of specialized circuits, and (2) greater divergence between longtail and head circuit behaviors exacerbates generalization failure. These insights directly inform our intervention strategies.

Building on our theoretical insights, we introduce Circuit-Aware Reward Training (CART), a methodology that identifies vulnerable circuits and guides targeted interventions. CART operates in three phases: circuit discovery, vulnerability assessment, and targeted intervention.

\subsection{Circuit Discovery}

The first stage is identifying which circuits specialize in longtail preference processing. We employ a multi-step approach:

\textbf{Activation Pattern Analysis}: We compute differential activation patterns between common and rare preference scenarios. For each neuron $j$, we calculate:
$$\Delta_j = \mathbb{E}_{x \sim p_{\text{tail}}}[a_j(x)] - \mathbb{E}_{x \sim p_{\text{head}}}[a_j(x)]$$

Neurons with high $|\Delta_j|$ values are candidates for longtail specialization.

\textbf{Circuit Coherence Analysis}: Individual neurons may participate in multiple circuits. We use mutual information to identify coherent groups of neurons that activate together for longtail inputs:
$$\text{Coh}(N_i, N_j) = I(N_i, N_j | x \sim p_{\text{tail}}) - I(N_i, N_j | x \sim p_{\text{head}})$$

High coherence values indicate neurons that form functional circuits for rare-event processing.

\textbf{Causal Validation}: Finally, we verify circuit functionality through activation patching: selectively modifying circuit activations and measuring the impact on reward predictions. This ensures that identified circuits are causally relevant rather than merely correlated.

\subsection{Vulnerability Assessment}

Once longtail circuits are identified, we assess their vulnerability to exploitation. This involves three complementary analyses:

\textbf{Prediction Consistency}: We measure how consistently circuits produce similar outputs for semantically similar longtail inputs. For a circuit $c \in \mathcal{C}_{\text{tail}}$, let $X_{\text{tail}}^{(c)} = \{x_1, \dots, x_m\}$ denote semantically similar longtail inputs activating $c$. The consistency score is
\begin{equation}
\text{Consist}(c) = 1 - \frac{1}{m} \sum_{i=1}^{m} \frac{|a_c(x_i) - \bar{a}_c|}{\max_j |a_c(x_j)|}, \quad
\bar{a}_c = \frac{1}{m} \sum_{i=1}^m a_c(x_i),
\end{equation}
where $a_c(x)$ is the activation of circuit $c$. High consistency indicates exploitable uncertainty.

\textbf{Adversarial Sensitivity}: We evaluate how small perturbations to longtail circuit activations affect final reward scores. Specifically,  sensitivity of the reward model to small perturbations in the circuit activation is measured by:
\begin{equation}
\text{Sens}(c) = \mathbb{E}_{x \sim p_{\text{tail}}} \left[ \max_{\|\delta\| \le \epsilon} \big| R_\theta(x; a_c + \delta) - R_\theta(x; a_c) \big| \right],
\end{equation}
where $R_\theta(x; a_c + \delta)$ denotes the reward when the activation of circuit $c$ is perturbed by $\delta$, bounded by $\epsilon$. High sensitivity indicates potential for reward hacking.

\textbf{Coverage Analysis}: We assess what fraction of the longtail distribution activates each circuit. Coverage measures the fraction of the longtail distribution effectively processed by circuit $c$:
\begin{equation}
\text{Cov}(c) = \Pr_{x \sim p_{\text{tail}}}\Big[ |a_c(x)| > \tau_{\text{act}} \Big],
\end{equation}
where $\tau_{\text{act}}$ is a threshold indicating significant activation. Circuits with low coverage create blind spots exploitable by policies.

\textbf{Composite Vulnerability Score.} We combine these metrics into a single vulnerability score:
\begin{equation}
\text{Vuln}(c) = \alpha \big(1-\text{Consist}(c)\big) + \beta \,\text{Sens}(c) + \gamma \big(1-\text{Cov}(c)\big),
\end{equation}
with hyperparameters $\alpha,\beta,\gamma$ weighting the contribution of each component. Circuits with high $\text{Vuln}(c)$ are prioritized for intervention. This allows identification of circuits most likely to contribute to reward hacking.

\subsection{Targeted Intervention}

Based on the vulnerability assessment, CART applies interventions to improve longtail robustness. 

\textbf{Circuit-Guided Data Augmentation.}  
Let $X_{\text{tail}}$ be the longtail dataset and $C_v \subseteq \mathcal{C}_{\text{tail}}$ the set of vulnerable circuits. We generate augmented examples $\tilde{x}$ that specifically activate circuits in $C_v$:
\begin{equation}
\tilde{x} = \arg\max_{x'} \sum_{c \in C_v} a_c(x'), \quad x' \sim \mathcal{G}(x),
\end{equation}
where $\mathcal{G}$ is a generative transformation of original data $x$. The augmented loss is:
\begin{equation}
\mathcal{L}_{\text{aug}}(\theta) = \frac{1}{|\tilde{X}|} \sum_{\tilde{x} \in \tilde{X}} \big(R_\theta(\tilde{x}) - y_{\tilde{x}}\big)^2,
\end{equation}
with $y_{\tilde{x}}$ the target reward for the augmented input.

\textbf{Circuit Regularization.}  
To stabilize vulnerable circuits, we introduce a variance-penalizing term:
\begin{equation}
\mathcal{L}_{\text{circuit}}(\theta) = \lambda \sum_{c \in C_v} \text{Var}_{x \sim p_{\text{tail}}}[a_c(x)],
\end{equation}
encouraging consistent activations across the longtail distribution.

\textbf{Progressive Circuit Strengthening.}  
We gradually emphasize longtail scenarios during training via a curriculum weight:
\begin{equation}
\mathcal{L}_{\text{prog}}(\theta) = \frac{1}{N} \sum_{i=1}^N w_t(x_i) \big(R_\theta(x_i) - y_i\big)^2, \quad
w_t(x) = \min\Big(1, \eta \cdot t \cdot \mathbb{1}[x \in X_{\text{tail}}]\Big),
\end{equation}
where $t$ is the training step and $\eta$ a scaling factor. Early in training, tail examples are downweighted to avoid overwhelming circuits; the weight increases over time.

\textbf{Circuit-Aware Ensembling.}  
To compensate for residual vulnerabilities, multiple models $\{R_\theta^{(k)}\}$ with complementary circuit specializations are combined:
\begin{equation}
R_{\text{ensemble}}(x) = \sum_{k=1}^K \alpha_k R_\theta^{(k)}(x), \quad \sum_k \alpha_k = 1,
\end{equation}
where $\alpha_k$ can be optimized to minimize longtail error:
\begin{equation}
\min_{\alpha} \frac{1}{|X_{\text{tail}}|} \sum_{x \in X_{\text{tail}}} \Big(R_{\text{ensemble}}(x) - y_x\Big)^2.
\end{equation}

\textbf{Combined Training Objective.}  
All components are integrated into a single objective:
\begin{equation}
\mathcal{L}_{\text{CART}}(\theta) = \mathcal{L}_{\text{head}}(\theta) + \mathcal{L}_{\text{aug}}(\theta) + \mathcal{L}_{\text{circuit}}(\theta) + \mathcal{L}_{\text{prog}}(\theta),
\end{equation}
where $\mathcal{L}_{\text{head}}$ is the standard reward loss on common scenarios. This objective provides a principled and mathematically grounded framework for improving longtail robustness.

\section{Discussion and Conclusion}

This work reframes reward hacking not merely as an optimization artifact but as a mechanistic failure rooted in longtail specialization. By analyzing the circuits that process rare inputs, we uncover how vulnerabilities arise and how they can be mitigated through targeted interventions. This perspective contributes explanatory power that goes beyond existing black-box defenses, while also suggesting concrete engineering strategies.

Our proposed Circuit-Aware Reward Training (CART) illustrates how mechanistic insights can be operationalized: identifying longtail circuits, diagnosing their vulnerabilities, and applying targeted training to improve robustness. Grounding alignment interventions in internal mechanisms provides a path toward more transparent and trustworthy reward models. In doing so, CART connects mechanistic interpretability with practical alignment practice, strengthening the bridge between scientific understanding and system design.

At the same time, important challenges remain. Mechanistic interpretability methods are still computationally intensive, raising questions about scalability to frontier-scale models. Circuits may not always exhibit clean specialization, as real networks often feature overlapping or dynamic structures. Evaluating robustness on longtail scenarios is itself difficult, requiring new benchmarks and protocols. Finally, while our framework was developed for language-model reward models, its extension to other modalities such as vision, robotics, or multimodal alignment remains an open question.

Addressing these challenges is crucial for realizing the full potential of circuit-based alignment strategies. We view this work as an initial step toward integrating interpretability and alignment at scale, with substantial room for methodological, theoretical, and practical development.

\bibliography{iclr2026_conference}

\end{document}